\documentclass{article} 
\usepackage{iclr2020_conference,times}

\usepackage{amsmath,amsfonts,bm}

\def\eqref#1{equation~\ref{#1}}

\def\1{\bm{1}}

\DeclareMathAlphabet{\mathsfit}{\encodingdefault}{\sfdefault}{m}{sl}
\SetMathAlphabet{\mathsfit}{bold}{\encodingdefault}{\sfdefault}{bx}{n}

\usepackage{hyperref}
\usepackage{url}
\usepackage{graphicx}
\usepackage{booktabs}
\usepackage{multirow}

\title{Towards Neural Machine Translation \\ for Edoid Languages}

\author{Iroro Fred \d{\`O}n\d{\`o}m\d{\`e} Orife \\
Niger-Volta Language Technologies Institute\\
\texttt{iroro@alumni.cmu.edu}
}

\iclrfinalcopy 
\begin{document}

\maketitle



\section{Introduction}

Many of the 500 plus languages spoken in Nigeria today have relinquished their previous prestige and purpose in modern society to English and Nigerian Pidgin, notably amongst the younger generations. Unlike numerous East and South Asian populations, which preserved the socio-linguistic status of their indigenous languages under colonial rule (e.g.\ India), Nigerian communities with primarily oral traditions have been the most susceptible to language endangerment \citep{rolle2013phonetics, omo2004esan}.

For tens of millions of speakers, language inequalities manifest themselves as unequal access to information, communications, health care, security along with attenuated participation in political and civic life. These inequities are further exacerbated in a technological age, where only the most highly resourced (i.e. colonial) languages become the milieu for economic advancement \citep{odojelanguage, awobuluyi201626, ganagana2019contrastive}. Finally, there have been practical and technical challenges in language technology for indigenous languages like orthographic standardizations and consistent diacritic representation (Unicode) in electronic media and across device types. 

For almost-extinct languages, machine translation offers hope for language documentation and preservation. For speakers of minority Nigerian languages, it can facilitate good governance, national development and offers a path for technological, economic, social and political participation and empowerment to those with unequal access \citep{odoje201612, odojelanguage}. Using the new JW300 public dataset, we trained and evaluated baseline Neural Machine Translation (NMT) models for four widely spoken Edoid languages: \d{\`E}d{\'o}, {\'E}s{\'a}n, Urhobo and Isoko. 


\section{Edoid Languages}


Belonging to the eastern sub-branch of the Volta-Niger family within the Niger-Congo phylum, and spoken by approximately 5 million people, the Edoid languages of Southern Nigeria (Edo and Delta states) comprise over two dozen so-called ``minority" languages. The term \emph{Edoid} stems from \d{\`E}d{\'o}, the most broadly spoken member langauge and the language of the famed Kingdom of Benin. \d{\`E}d{\'o}, {\'E}s{\'a}n are members of the North-Central branch while Urhobo and Isoko belong to the South-Western family \citep{ethnologue_2019}. These languages were selected based on the availability of text and because they are the most widely spoken.

Edoid langauges generally employ the SVO constituent order type, open syllable systems with very few consonant clusters. Each language has at least two basic tone levels, high (H) and low (L) with kinetic, downstepped or contour tones variously utilized. As tone patterns serve different lexical and grammatical functions, ``the phonetic and phonological implementation of this system is in fact complex and difficult to pin down" \citep{rolle2013phonetics, ogie2009multi, adeniyi2010tone, ilolo2013vowel}. Finally, nasalisation is very common for both vowels and consonants \citep{Elugbe_1989, isoko_phonetics, ikoyo2018phonetic}. 


Within Nigeria there is scholarship on rule, phrase and statistical machine translation systems for majority tongues of Yor{\`u}b{\'a}, Igbo and Hausa \citep{odojelanguage}. The present study is the first work known to the authors done in computational linguistics for any of the Edoid langauges, specifically for machine translation.

\section{Methodology}
\label{methods}

We first built baseline models using the Transformer architecture, the dominant modeling approach for NMT. The Transformer uses an encoder-decoder structure with stacked multi-head self-attention and fully connected layers \citep{NIPS2017_7181}. Given the performance of Byte Pair Encoding (BPE) subword tokenization for low-resourced South African languages, and the size of our datasets, we trained baseline models based on the ablation study results by Martinus et al., some 4000 BPE tokens \citep{focus_southafrica}. Models were then re-trained for all four languages using the standard word-level tokenization.

\paragraph{Dataset:} The recently published JW300 dataset is a large-scale, parallel corpus for Machine Translation (MT) comprising more than three hundred languages of which 101 are African \citep{agic-vulic-2019-jw300}. JW300 text is drawn from the Watchtower and Awake! religious magazines by Jehovah's Witnesses (JW). The test set contains sentences with the highest coverage across all other languages in the corpus. The cardinality of the training set in number of tokens and sentences is listed in Appendix Table~\ref{results}. 

\paragraph{Models:} The open-source, Python 3 machine translation toolkit \texttt{JoeyNMT} was used to train Transformer models \citep{JoeyNMT}. Our training hardware was the free-tier configuration on Google Colaboratory, a single core Xeon CPU instance and a Tesla K80 GPU. Model training elapsed over multiple days, as experiments were repeated for the different tokenizations.

\section{Results}
\label{results}

\paragraph{Qualitative:} Urhobo and Isoko with larger training texts unsurprisingly had higher BLEU scores which generally correlated with the translation quality when reviewed by L1 speakers. BPE tokenization provided approximately a 37\% boost across dev and test sets for \d{\`E}d{\'o} and {\'E}s{\'a}n, a 32\% boost for Urhobo but was flat to slightly worse than word-level tokenization for Isoko. Full scores and examples are listed in the Appendix.

\begin{table}[h]
\caption{Per-language BLEU scores by BPE or word-level tokenization}
\label{results}
\begin{center}
\begin{tabular}{c@{\qquad}ccc@{\qquad}ccc}
  \toprule
  \multirow{2}{*}{\raisebox{-\heavyrulewidth}{\textbf{Language}}} & \multicolumn{2}{c}{\textbf{BPE}} & \multicolumn{2}{c}{\textbf{Word}} & \multirow{2}{*}{\raisebox{-\heavyrulewidth}{\textbf{Tokens}}} & \multirow{2}{*}{\raisebox{-\heavyrulewidth}{\textbf{Sentences}}}
  	 \\
  \cmidrule{2-5}
  & dev & test & dev & test \\
  \midrule
  \d{\`E}d{\'o}  & 7.92 & 12.49 & 5.99 & 8.24 &  229,307 & 10,188 \\
  {\'E}s{\'a}n & 4.94 & 6.25 & 3.39 & 5.30 & 87,025 & 4,128 \\
    \midrule
  Urhobo  & 15.91 & 28.82 & 11.80 & 22.39 & 519,981 & 25,610 \\
  Isoko   & 32.58 & 38.05 & 32.38 & 38.91 & 4,824,998 & 214,546 \\
  \bottomrule
  \end{tabular}
\end{center}
\end{table}

\paragraph{Error Analysis:} While studying the models' predictions, we observed that the training data requires significantly more preprocessing, notably to remove prevalent scriptural book, chapter and verse number annotations. Cleaner data will simplify the training task and generate models which generalize better on non-Biblical texts. Dialects of a singular language can also exhibit variance in the expression of concepts, so it would be advantageous to capture multiple references in different dialects. Finally, based on the performance of the Isoko models, with BLEU scores in the range \{32, 39\}, we have an estimate of how much additional clean text is required to achieve a similar performance with \d{\`E}d{\'o} and {\'E}s{\'a}n.

\section{Future Work and Conclusions}
Fertile avenues for future work include investigations using unsupervised machine translation and back-translation, a full ablation study with different (subword) tokenization approaches as well as specific consideration of linguistic knowledge. A human evaluation study accompanied by a more extensive error analysis will also be crucial to better understand the linguistic features where NMT models under-perform.

We hope this initial effort will assist translators and the lay-person alike, bootstrap development and sustenance of literary traditions and energize interest in language technology for socio-linguistic and economic empowerment. Ultimately, languages with predominantly oral traditions will benefit most from (audio) speech-to-speech language technologies \citep{jia2019direct}. This present work is but one step towards that goal. All public-domain datasets, pre-trained translation models and their training and evaluation configurations are available on GitHub.\footnote{\url{https://github.com/Niger-Volta-LTI/edoid-nmt}}
 
%



\subsubsection*{Acknowledgments}
The authors thank Dr. Ajovi B. Scott-Emuakpor, MD and Dr. John Nevboyeri Orife for their encouragement and qualitative critiques of the translations.

\bibliography{iclr2020_conference}
\bibliographystyle{iclr2020_conference}

\appendix
\section{Appendix}

\begin{table}[h]
\caption{Example Translations}
\label{translations}
\begin{center}
  \begin{tabular}{ll}
     \textbf{\d{\`E}d{\'o}}  & \\
     \midrule
     \midrule
     Source:   &  Reading and meditating on real - life Bible accounts can help us to do what ?  \\
Reference: & De vbene okha ni rre Baibol ya ru iyob\d{o} ne ima h\d{e} ?  \\
Hypothesis: & De emwi ne ima gha ru ne ima mieke na gha mw\d{e} ir\d{e}nmwi n\d{o} gbae vbekpae Jehova ?  \\
     \midrule
	Source:   &    What are the rewards for being humble ? \\
	Reference:  &  Ma ghaa mu egbe rriot\d{o} , de afiangbe na lae mi\d{e}n ? \\
	Prediction:  & De emwi n\d{o} kh\d{e}ke ne \d{o}mwa n\d{o} dizigha \d{o}y\d{e}vbu ru ? \\
	 \bottomrule
     \\
    \textbf{{\'E}s{\'a}n}  & \\
     \midrule
     \midrule
	Source:      &  I WAS raised in Graz , Austria . \\
	Reference:    & AGBA\d{E}BHO nati\d{o}le Graz bhi Austria , \d{o}le m\d{e}n da wanre . \\
	Prediction:   & M\d{e}n da ha khian \d{o}ne isikulu , m\d{e}n da d\d{o} ha khian \d{o}ne isikulu . \\
     \midrule
     Source:    &   We should also strive to help others spiritually . \\
	 Reference:  &  Ahami\d{e}n mhan re \d{e}ghe bhi ot\d{o} r\d{e} ha lu\d{e} iBaibo  \\
	 Prediction:  &  Mhan d\d{e} sab\d{o} r\d{e}kpa mhan r\d{e} sab\d{o} ha mh\d{o}n ur\d{e}\d{o}bh\d{o} b\d{o}si eria . \\
	 \bottomrule
	\\
    \textbf{Urhobo}  & \\
     \midrule
     \midrule
	Source:    &                   But freedom from what ? \\
	Reference: &  		   \d{E}k\d{e}vu\d{o}vo , \d{e}dia v\d{o} yen egbom\d{o}ph\d{e} na che si ayen nu ? \\
	Prediction: & ( 1 Pita 3 : 1 ) \d{E}k\d{e}vu\d{o}vo , die yen egbom\d{o}ph\d{e}  \\
     \midrule

	Source:    &   Today he is serving at Bethel . \\ 
	Reference:  &  Non\d{e}na , \d{o} ga vw\d{e} B\d{e}t\d{e}l .\\
	Prediction: &  Non\d{e}na , \d{o} ga vw\d{e} B\d{e}t\d{e}l asa\d{o}kiephana . \\
	\bottomrule
	\\
    \textbf{Isoko}  & \\
     \midrule
     \midrule
	Source:      & Still , words of apology are a strong force toward making peace . \\
	Reference:   & Ghele na , eme unu - uwou u re fi ob\d{o} h\d{o} gaga eva\d{o} eruo udhedh\d{e} .\\
	Prediction:  & Ghele na , eme unu - uwou y\d{o} \d{e}gba ologbo n\d{o} ma re ro ru udhedh\d{e} .\\
  \midrule
	 Source:   & We can even ask God to ` create in us a pure heart . ' \\
	Reference:  & Ma r\d{e} sae tub\d{e} yare \d{O}gh\d{e}n\d{e} re \d{o}  ` k\d{e} omai eva efuafo . ' \\
	Prediction: & Ma r\d{e} sae tub\d{e} yare \d{O}gh\d{e}n\d{e} re \d{o}  ` ma omai eva efuafo .  \\
	 \bottomrule
     
  \end{tabular}
\end{center}
\end{table}

\end{document}